\documentclass[10pt,twocolumn,letterpaper]{article}

\usepackage[pagenumbers]{cvpr} 

\usepackage[dvipsnames]{xcolor}

\usepackage{multirow}
\usepackage{times}
\usepackage{epsfig}
\usepackage{amsfonts}
\usepackage{array}
\usepackage{caption}
\usepackage{balance}
\usepackage{booktabs}
\usepackage{tikz}
\usepackage{pifont}
\usepackage{mathrsfs}
\usepackage{xspace}
\usepackage{sidecap}
\usepackage{mathtools}
\usepackage{xcolor,colortbl}
\usepackage{indentfirst}

\usepackage[pagebackref,breaklinks,colorlinks,citecolor=cvprblue]{hyperref}
\usepackage{chngcntr}
\newcommand{\cmark}{\ding{51}}

\definecolor{cvprblue}{rgb}{0.21,0.49,0.74}

\newcommand{\pc}{PC$^{2}$\xspace}
\newcommand{\ccd}{CCD-3DR\xspace}
\newcommand{\down}{$\downarrow$}
\newcommand{\up}{$\uparrow$}
\newcommand{\bdm}{BDM\xspace}

\newcommand{\bdmx}{{\bf x}}
\newcommand{\bdmy}{{\bf y}}

\def\paperID{13157} 
\def\confName{CVPR}
\def\confYear{2024}

\definecolor{lightgray}{HTML}{f2f2f2}

\newcommand{\cellcolorlightgray}{\cellcolor{lightgray}}

\title{Bayesian Diffusion Models for 3D Shape Reconstruction}

\author{%
  Haiyang Xu$^{*,1}$ \qquad Yu Lei$^{*,2}$ \qquad Zeyuan Chen$^{3}$ \qquad
  Xiang Zhang$^{3}$ \\
  Yue Zhao$^{4}$ \qquad Yilin Wang$^{4}$ \qquad
  Zhuowen Tu$^{3}$ \\
  $^1$ University of Science and Technology of China \quad $^2$ Shanghai Jiao Tong University \\
  $^3$ University of California, San Diego \quad $^4$ Tsinghua University
}

\begin{document}
\maketitle

\begin{abstract}

\vspace{-0.5em}

We present Bayesian Diffusion Models (BDM), a prediction algorithm that performs effective Bayesian inference by tightly coupling the top-down (prior) information with the bottom-up (data-driven) procedure via joint diffusion processes. We show the effectiveness of BDM on the 3D shape reconstruction task. Compared to prototypical deep learning data-driven approaches trained on paired (supervised) data-labels (\textit{e.g.} image-point clouds) datasets, our BDM brings in rich prior information from standalone labels (\textit{e.g.} point clouds) to improve the bottom-up 3D reconstruction. As opposed to the standard Bayesian frameworks where explicit prior and likelihood are required for the inference, BDM performs seamless information fusion via coupled diffusion processes with learned gradient computation networks. The specialty of our BDM lies in its capability to engage the active and effective information exchange and fusion of the top-down and bottom-up processes where each itself is a diffusion process. We demonstrate state-of-the-art results on both synthetic and real-world benchmarks for 3D shape reconstruction. \textit{Project link: \href{https://mlpc-ucsd.github.io/BDM}{https://mlpc-ucsd.github.io/BDM}}
\let\thefootnote\relax\footnotetext{* equal contribution. Work done during the internship of Haiyang Xu, Yu Lei, Yue Zhao, and Yilin Wang at UC San Diego.}

\vspace{-1.75em}

\end{abstract}    
\section{Introduction \label{sect:intro}}

The Bayesian theory \cite{bernardo2009bayesian,knill1996perception}, under the general principle of \textit{analysis-by-synthesis} \cite{yuille2006vision}, has made a profound impact on a myriad of tasks in computer vision and machine learning, including face modeling \cite{cootes1995active}, shape detection and tracking \cite{blake2012active}, image segmentation \cite{tu2002image}, scene categorization \cite{fei2005bayesian}, image parsing \cite{tu2005image}, depth estimation \cite{liu2010single,woodford2009global}, object recognition \cite{weber2000unsupervised,felzenszwalb2005pictorial,fergus2003object}, and topic modeling \cite{blei2003latent}. 

We assume the task of predicting $\bdmy$ for a given input $\bdmx$ ($\bdmy$ and $\bdmx$ represent respectively the 3D point clouds and the input image in this paper). The Bayes' theorem turns the posterior $p(\bdmy|\bdmx)$ into the product of the likelihood $p(\bdmx|\bdmy)$ and the prior $p(\bdmy)$ as $p(\bdmy|\bdmx) \propto p(\bdmx|\bdmy) p(\bdmy)$, which can be further approximated by $p_{\gamma}(\bdmy|\bdmx) p(\bdmy)$ \cite{lafferty2001conditional}, where $p_{\gamma}(\bdmy|\bdmx)$ represents a direct bottom-up (data-driven) process, \textit{e.g.}, the Viola-Jones face detector \cite{viola2001rapid}. The formulations of $p(\bdmx|\bdmy) p(\bdmy)$ \cite{li2009markov} and $p_{\gamma}(\bdmy|\bdmx) p(\bdmy)$ \cite{lafferty2001conditional} can be solved via \textit{e.g.}, the Markov Chain Monte Carlo (MCMC) sampling methods \cite{andrieu2003introduction,welling2011bayesian}.

\begin{figure}[!tp]
\centering
\includegraphics[width=\linewidth]{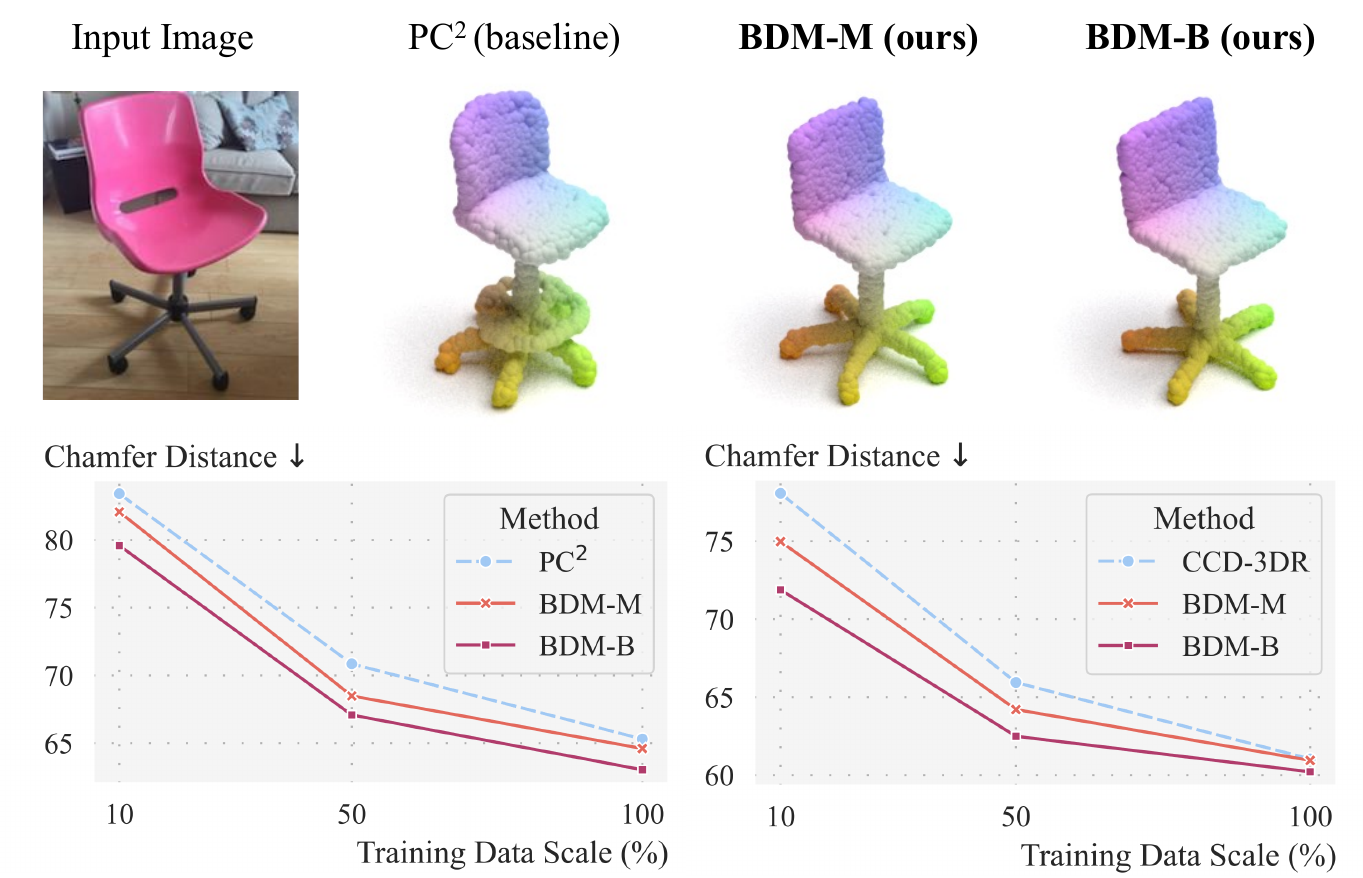}
\caption{\textbf{Baseline vs Bayesian Diffusion Models}. Our BDM brings rich prior knowledge into the shape reconstruction process, fixing the incorrect predictions by the baseline (top row). BDM surpasses baselines in all three training data scales (bottom row).}
\label{fig:teaser}
\vspace{-1.25em}
\end{figure}

\vspace{0.2em}
\noindent \textbf{When does the Bayesian help?} The Bayesian theory \cite{bernardo2009bayesian,yuille2006vision,jordan1994hierarchical} provides a principled statistical foundation to guide the top-down and bottom-up inference, which is deemed to be of great biological significance \cite{knill1996perception}. In the early development of computer vision \cite{marr2010vision}, the objects in study \cite{cootes1995active,fergus2003object,weber2000unsupervised} are often of simplicity and are picked from relatively small-scale datasets \cite{fei2007learning,griffin2007caltech,nilsback2008automated}. The top-down prior \cite{tu2005image,felzenszwalb2005pictorial} can therefore provide a strong regularization and inductive bias to the bottom-up process that has been trained from the data \cite{fei2007learning,griffin2007caltech}.

\vspace{0.2em}
\noindent \textbf{Why is the Bayesian not anymore widely adopted in the deep learning era?} Although still being an active subject \cite{fortuin2022priors} in study, the top-down/prior information has not been widely adopted in the big-data/deep-learning era, where an immediate improvement can be shown over the data-driven models \cite{krizhevsky2012imagenet,dosovitskiy2020image,ren2015faster} learned from large-scale training set of input and ground-truth pairs $S_s=\{(\bdmx_i, \bdmy_i), i=1..n \}$. The reasons are threefold: \textbf{1)} \textit{Rich features} from large-scale data \cite{deng2009imagenet} become substantially more robust than manually designed ones \cite{lowe2004distinctive}, whereas the top-down prior $p(\bdmy)$ for structured output $\bdmy$ no longer shows an improvement. \textbf{2)} \textit{Strong bottom-up models} \cite{krizhevsky2012imagenet,dosovitskiy2020image,ren2015faster} learned in a $\bdmx \to \bdmy$ fashion from paired/supervised dataset, $S_s=\{(\bdmx_i, \bdmy_i), i=1..n \}$, are powerful, and they do not necessarily see an immediate benefit from introducing a separate prior $p(\bdmy)$ that is obtained from $S_l=\{\bdmy_i, i=1..n \}$ alone, as knowledge about the $\bdmy$ has already been implicitly captured in the data-driven $p_{\gamma}(\bdmy|\bdmx)$. \textbf{3)} The presence of the intermediate stages with different architectural designs for the deep models makes merely combining the data-driven model $p_{\gamma}(\bdmy|\bdmx)$ and the prior $p(\bdmy)$ not so obvious, as the distributions for $p_{\gamma}(\bdmy|\bdmx)$ and $p(\bdmy)$ are hard to model and may not co-exist.

\vspace{0.2em}
\noindent \textbf{The emerging opportunity for combining bottom-up and top-down processes with diffusion-based models.} The recent development in diffusion models \cite{sohl2015deep, ho2020ddpm,song2020score,rombach2022high} has led to substantial improvements to unsupervised learning beyond the traditional VAE \cite{kingma2013auto} and adversarial learning \cite{tu2007learning,goodfellow2014generative,jin2017introspective}. The presence of diffusion models for learning both $p(\bdmy)$ (\textit{e.g.} shape priors \cite{zhou2021pvd}) and $p_{\gamma}(\bdmy|\bdmx)$ (\textit{e.g.} 3D shape reconstruction \cite{melas2023pc2,di2023ccd}) inspires us to develop a new inference algorithm, Bayesian Diffusion Models, that is applied to single-view 3D shape reconstruction.

\vspace{0.2em}
The contribution of our paper is summarized as follows:

\begin{itemize}
    \item We present \textbf{Bayesian Diffusion Models} (BDM), a new statistical inference algorithm that couples diffusion-based bottom-up and top-down processes in a joint framework. BDM is particularly effective when having separately available data-labeling (supervised) dataset $S_s=\{(\bdmx_i, \bdmy_i), i=1..n \}$ and standalone label dataset $S_l=\{\bdmy_i, i=1..m \}$ for training $p_{\gamma}(\bdmy|\bdmx)$ and $p(\bdmy)$ respectively. For example, obtaining a set \cite{pix3d} for real-world images with the corresponding ground-truth 3D shapes is challenging whereas a large dataset of standalone 3D object shapes such as ShapeNet \cite{chang2015shapenet} is readily available.
    \item Two strategies for fusing the information exchange between the bottom-up and the top-down diffusion process are developed: 1) a \textbf{blending procedure} that takes the two processes in a plug-in-and-play fashion, and 2) a \textbf{merging procedure} that is trained.
    \item We emphasize the key property of \textbf{fusion-with-diffusion} in BDM vs. \textbf{fusion-by-combination} in the traditional MCMC Bayesian inference. BDM also differs from the current pre-training + fine-tuning process \cite{brown2020language} and the prompt engineering practice \cite{ho2022cfg} by making the bottom-up and top-down integration process transparent and explicit; BDM points to a promising direction in computer vision and machine learning with a new diffusion-based Bayesian method.
\end{itemize}

BDM demonstrates the state-of-the-art results on the single image 3D shape reconstruction benchmarks.

\section{Related Work}

\begin{figure*}[htbp]
\centering
\includegraphics[width=\linewidth,trim=1.5em 0 1.5em 0,clip]{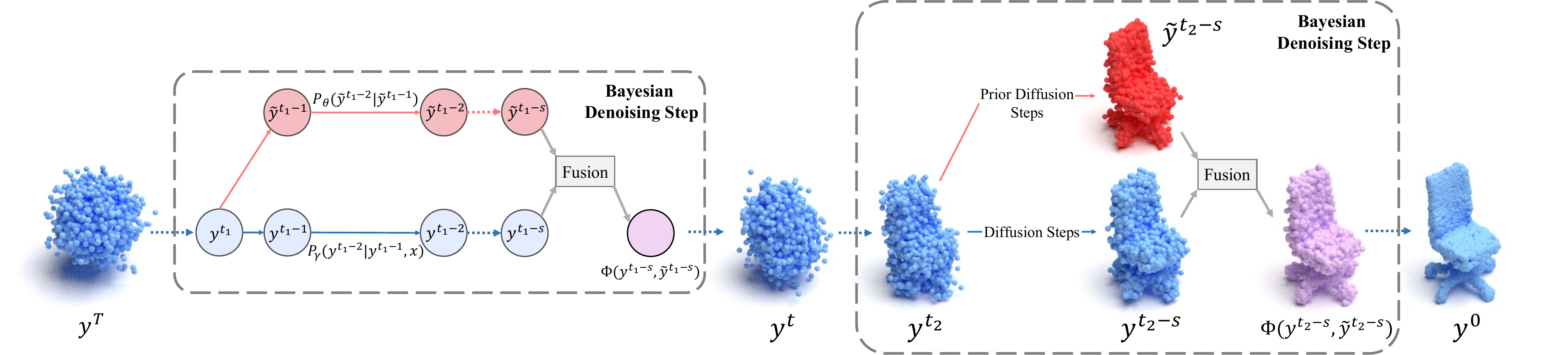}
\caption{Overview of the generative process in our Bayesian Diffusion Model. In each Bayesian denoising dtep, the prior diffusion model fuses with the reconstruction process, bringing rich prior knowledge and improving the quality of the reconstructed point cloud. We illustrate our Bayesian denoising step in two ways, left in the form of a flowchart and right in the form of point clouds.}
\label{fig:overview_pointcloud}
\vspace{-1em}
\end{figure*}

\noindent \textbf{Bayesian Inference.} As stated previously, 
the Bayesian theory \cite{bernardo2009bayesian,knill1996perception} has been adopted in a wide range of computer applications\cite{cootes1995active,blake2012active,fei2005bayesian,tu2005image,liu2010single,woodford2009global,weber2000unsupervised,felzenszwalb2005pictorial,fergus2003object}, but the results of these approaches are less competitive than those by the deep learning based ones \cite{krizhevsky2012imagenet,he2016deep,ren2015faster}. 

\vspace{0.3em}
\noindent \textbf{3D Shape Generation.} 
Early 3D shape generation methods \cite{gadelha20173d,henzler2019escaping,wu2016learning,achlioptas2018learning,yang2019pointflow,park2019deepsdf} typically leverage variational auto-encoders (VAE) \cite{kingma2013auto} and generative adversarial networks (GAN) \cite{goodfellow2014generative} to learn the distribution of the 3D shape. Recently, the superior performance of diffusion models in generative tasks also makes them the go-to methods in 3D shape generation. PVD \cite{zhou2021pvd} proposes to diffuse and denoise on point clouds using a Point-Voxel-CNN \cite{liu2019point}, while in DMPGen \cite{luo2021diffusion}, the diffusion process is modeled by a PointNet \cite{qi2017pointnet}. LION \cite{zeng2022lion} uses a hierarchical VAE to encode 3D shapes into latents where the diffusion and generative processes are performed. Another popular category of 3D generative models leverages 2D text-to-image diffusion models as priors and lifts them to 3D representations \cite{poole2022dreamfusion, lin2023magic3d,wang2023prolificdreamer,nichol2022point}.

\vspace{0.3em}
\noindent \textbf{Single-View 3D Reconstruction.}
Recovering 3D object shapes from a single view is an ill-posed problem in computer vision. Traditional approaches extract multi-modal information, including shading \cite{atick1996statistical, horn1970shape}, texture \cite{witkin1981recovering}, and silhouettes \cite{cheung2003visual}, for reconstructing 3D shapes. Learning-based reconstruction methods become popular with the advance of neural networks and the availability of large-scale 2D-3D datasets \cite{chang2015shapenet,fu20213d}. In these methods, different 3D representations are employed, including voxel grids \cite{wu20153d,girdhar2016learning,choy2016r2n2,xie2019pix2vox,zhang2023uni}, point clouds \cite{fan2017point,mandikal2019dense}, meshes \cite{kar2015category,kanazawa2018learning,wu2022casa}, and implicit functions \cite{mescheder2019occupancy,saito2019pifu,chibane2020implicit}. 

Some other reconstruction methods take images as conditions for generative models \cite{wu2016learning,melas2023pc2,di2023ccd}. In particular, they learn the prior shape distribution from large-scale 3D datasets and then perform 3D reconstruction with 2D image observations. 3D-VAE-GAN \cite{wu2016learning} employs GAN and VAE to learn a generator mapping from a low-dimensional probabilistic space to a 3D-shape space, and feeds images to the generator for reconstruction. Recent methods for single-view reconstruction are mostly based on diffusion models \cite{melas2023pc2,di2023ccd,liu2023zero,liu2023one}. \pc
\cite{melas2023pc2} proposes to project encoded features from 2D back to 3D, which facilitates the point cloud reconstruction in denoising. \ccd \cite{di2023ccd} introduces a centered diffusion probabilistic model based on \pc, which offers better consistency in alignments of local features and final prediction results. RenderDiffusion \cite{anciukevivcius2023renderdiffusion} presents an explicit latent 3D representation into a diffusion model, yielding a 3D-aware pipeline that could perform 3D reconstruction. Zero1-to-3 \cite{liu2023zero} and One-2-3-45 \cite{liu2023one} inject camera information into a 2D diffusion model and reconstruct 3D shapes with synthesized multi-view images.

\vspace{0.3em}
\noindent \textbf{Prompts and Latent Representations in Transformers.}
Transformers \cite{vaswani2017attention} provide a general tokenized learning framework, allowing the insertion of the representation of $\bdmy$ into $p_{\gamma}(\bdmy|\bdmx)$ as special tokens \cite{ho2022cfg,cheung2003visual}. However, the prior knowledge in Transformers serves as a latent condition that is often opaque and non-interpretable.

\vspace{-0.5em}

\section{Method}

In the following section, we introduce the Bayesian Diffusion Models, the framework of which is illustrated in \cref{fig:overview_pointcloud}. Initially, a concise overview of denoising diffusion models, particularly focusing on point cloud diffusion models, is presented. This is followed by an exposition of its conditional variant applied to 3D shape reconstruction. Subsequently, we delve into the central concept of our proposed Bayesian Diffusion Models which integrate Bayesian priors. Concluding this section, we provide an in-depth exploration of our novel prior-integration methodology.

\begin{figure*}[!htp]
\centering
\includegraphics[width=\linewidth, trim=0 1em 0 1em, clip]{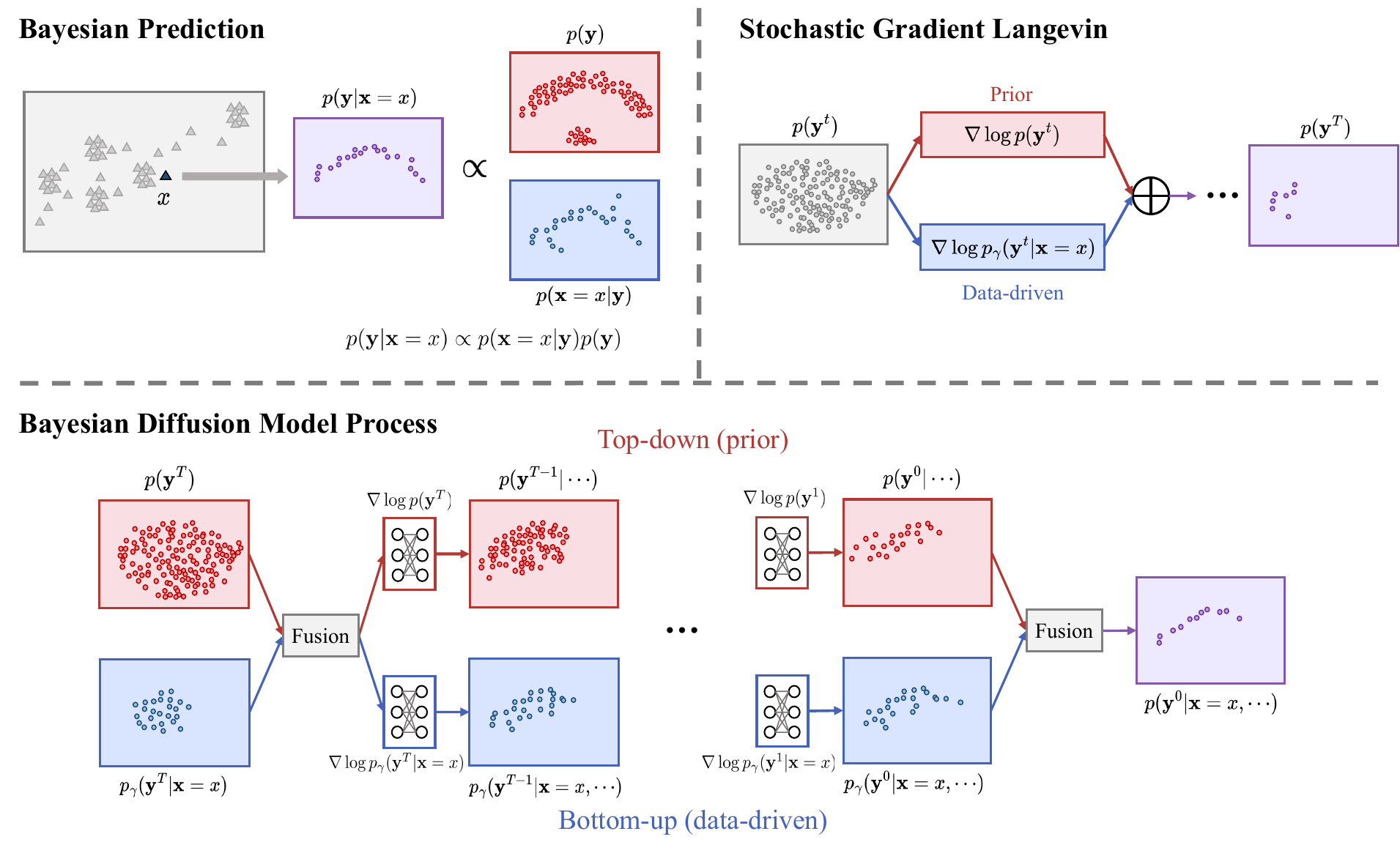}
\caption{Illustration for the Bayesian Diffusion Models compared with the standard Bayesian formulation. We present the standard Bayesian formulation and the one using stochastic gradient Langevin on the top part, while our proposed BDM on the bottom.}
\label{fig:bayes}
\vspace{-1em}
\end{figure*}

\subsection{Bayesian Inference with Stochastic Gradient Langevin Dynamics}

As discussed in \cref{sect:intro}, our task is to predict $\bdmy$ (a set of point clouds) for a given input $\bdmx \in \mathbb{R}^q$ (an input image). For the 3D shape reconstruction task, the output $\bdmy$ consists of a set of 3D points. The Bayesian theory \cite{bernardo2009bayesian} focuses on the study of the posterior $p(\bdmy|\bdmx) \propto p(\bdmx|\bdmy) p(\bdmy)$, where $p(\bdmy)$ is the prior (top-down information). The likelihood term $p(\bdmx|\bdmy)$ can be alternatively replaced \cite{lafferty2001conditional} by a data-driven (bottom-up) distribution $p_{\gamma}(\bdmy|\bdmx)$, which is learned from a paired data-labels training set $S_s=\{(\bdmx_i, \bdmy_i), i=1..n \}$. Therefore, the inference for the optimal prediction $\bdmy^*$ can then be carried via Markov Chain Monte Carlo (MCMC) \cite{andrieu2003introduction} using stochastic gradient Langevin dynamics \cite{welling2011bayesian}:
\begin{eqnarray}  
    & & \Delta \bdmy^t = \frac{\epsilon^t}{2} \big ( \underbrace{\nabla \log p_{\gamma}(\bdmy^t|\bdmx)}_{data-driven} + \underbrace{\nabla \log p(\bdmy^t)}_{prior} \big ) + \eta^t \nonumber \\
    & & \eta^t \sim N(0, \epsilon^t)
    \label{eqn:bayes_langevin}
\end{eqnarray}
where $\epsilon^t$ denotes a sequence of step size and $\eta^t$ is a Gaussian noise.  The challenge in implementing Equation~\ref{eqn:bayes_langevin} is twofold: \textbf{1)} it requires knowing the explicit formulation for both the $\log p_{\gamma}(\bdmy^t|\bdmx)$ and $\log p(\bdmy^t)$,  which is hard to obtain in real-world applications. \textbf{2)} a mere summation for the gradients $\nabla \log p_{\gamma}(\bdmy|\bdmx)$ and $\nabla \log p(\bdmy^t)$ limits the level of interaction between the bottom-up and top-down processes. \cref{fig:bayes} shows the basic Bayes formulation and demonstrates the stochastic gradient Langevin inference result. 

\subsection{Denoising Diffusion Probabilistic Models}

The denoising diffusion models \cite{sohl2015deep, ho2020ddpm,song2020score} have demonstrated superior performance in representing the structured data of high-dimension in both paired data-labels setting (learning $\nabla \log p^t(\bdmy)$ from $S_l=\{\bdmy_i, i=1..m \}$ \cite{zhou2021pvd}) and standalone labels settings (learning $\nabla \log p_{\gamma}^t(\bdmy|\bdmx)$ from $S_s=\{(\bdmx_i, \bdmy_i), i=1..n \}$ \cite{melas2023pc2}) manners. 
 
Here, we discuss the general formulation of the DDPM model \cite{sohl2015deep, ho2020ddpm} in the context of single image 3D shape reconstruction.
For point cloud generation, diffusion models are adept at learning the 3D shape of objects represented in point cloud format. At a high level, diffusion models iteratively denoise 3D points from a Gaussian sphere into a recognizable object. Consider a set of point clouds ${\bdmy^0}$, consisting of $N$ points, as an object in a $3N$-dimensional space. The diffusion model is employed to learn the mapping $s_\theta: \mathbb{R}^{3N} \rightarrow \mathbb{R}^{3N}$. Specifically, it is designed to estimate $q(\bdmy^{t-1} | \bdmy^{t})$, which is the offset of the points from its position at timestep $t$. This approach aligns with standard diffusion model practices, which are trained to predict the noise $\epsilon \in \mathbb{R}^{3N}$ using the following loss function:
\begin{equation}
    \mathcal{L} = \mathbb{E}_{\epsilon \sim \mathcal{N}(0, \mathbf{I})}\left[\left\|\epsilon - s_\theta(\bdmy^t, t)\right\|_2^2\right].
\end{equation}

The process of single-view reconstruction can be considered as a conditional point cloud generation. Diffusion models can be effectively applied here as well. Given a single-view image $\bdmx$, the diffusion model aims to estimate $q(\bdmy_{t-1} | \bdmy_{t}, \bdmx)$. Therefore, we should sample from $q(\bdmy_{t-1} | \bdmy_{t}, \bdmx)$ for this task. All the other concepts and principles are the same as point cloud generation. 

\subsection{Bayesian Diffusion Model}

In our work, we denote $\bdmx$ as an input image instance, while $\bdmy^t, t= 1 \cdots T $ to represent the 3D object we want to reconstruct. $\gamma$ is taken as the distribution learned by the reconstruction model and $\Pi$ represents the distribution after fusing prior and $\gamma$.

\noindent \textbf{Generative model:} The shapes generated by prior model can be formulated as $p\left(\bdmy^0\right)$. The noise predicted by the model at every timestep drives the spatial distribution according to the Markov Chain state transition probability $p\left(\bdmy^{t} \mid \bdmy^{t+1}\right)$. For the gradient, we have:
\begin{equation}
\varepsilon \left(\bdmy^t\right) = -\sigma_t \nabla_{\bdmy^t} \log p(\bdmy^t )
\end{equation}
 
\noindent \textbf{Reconstruction model:} In the same way of understanding generative diffusion model, the shapes generated by $\gamma$ can be formulated as $p_{\gamma}\left(\bdmy^0 \mid \bdmx \right)$. 
The model can reconstruct point clouds from learned $p_{\gamma}\left(\bdmy^t \mid \bdmy^{t+1}, \bdmx \right)$. In a similar way to the above, we have:
\begin{equation}
\varepsilon_{\gamma} \left( \bdmy^t, \bdmx \right) = -\sigma_t \nabla_{\bdmy^t} \log p_{\gamma}(\bdmy^t \mid \bdmx)
\end{equation}

It is obvious to conclude that the state transition probability of our Bayesian Diffusion Model comes from the corresponding probability from model $\gamma$ and posterior from the prior model. As show in \cref{fig:overview_pointcloud}, we use an inaccessible function $\mathbf{\Phi}$ to incorporate the prior diffusion gradient into the reconstruction diffusion gradient. Accordingly, we can deduce the fused gradient in our Bayesian Diffusion Model as follows:
\begin{equation}
\resizebox{.9\linewidth}{!}{
$
\begin{aligned}
\nabla \log p_\pi(\bdmy^t \mid \bdmx)  
& = \nabla_\gamma \log \Phi \big ( p_\gamma(\bdmy^t \mid \bdmx, \tilde{\bdmy}^{t+1} \sim p_\pi(y^{t+1} \mid \bdmx), \\
& p(\bdmy^t \mid \tilde{\bdmy}^{t+1} \sim p_\pi(\bdmy^{t+1} \mid \bdmx) \big )
\end{aligned}
$
}
\label{eqn:bdm_fusion}
\end{equation}
\subsection{Point Cloud Prior Integration}

As stated in \cref{sect:intro}, diffusion-based models have provided us with the possibility to benefit from both bottom-up and top-down processes. Specifically, the multi step inference procedure allows more flexible and effective forms of function $\mathbf{\Phi}$ in \cref{eqn:bdm_fusion}. Therefore, we employ step-wise interaction between a generative diffusion model and a reconstruction model, facilitating closer integration of point cloud priors. In particular, we feed the intermediate point cloud from our reconstruction model into the prior model, forward it through a certain number of timesteps in both models and fuse two new point clouds from the prior model and the reconstruction model as the input for the reconstruction model in the next time step. As below we introduce two fusion methods: BDM-M (Merging) and BDM-B (Blending). Both are carried out under fusion-with-diffusion.

\begin{figure}[!ht]
    \centering
    \includegraphics[width=\linewidth]{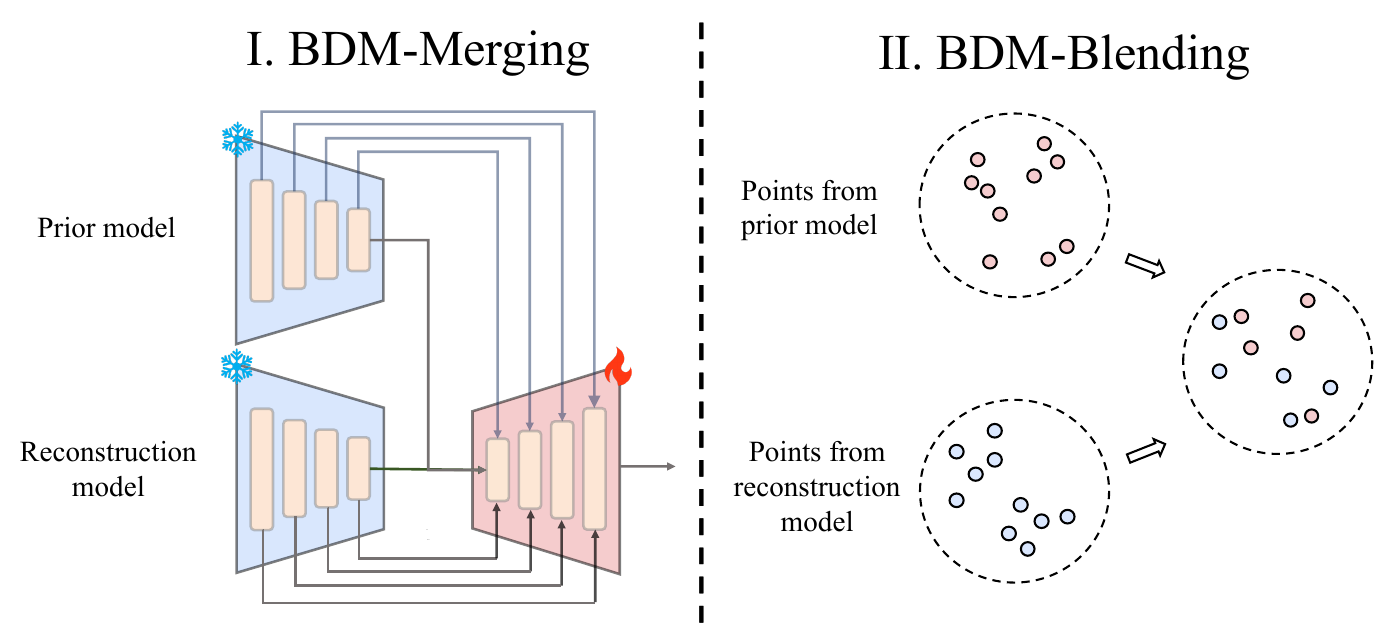}
    \caption{Illustration of our proposed fusion methods: BDM-M and BDM-B. The left part is the BDM-M, while the right side shows the BDM-B.}
    \label{fig:fusion}
    \vspace{-1em}
\end{figure}

\vspace{-1em}

\subsubsection{BDM-M (Merging)} In Merging, we propose a learnable paradigm for incorporating knowledge from the prior model into the reconstruction model. We implement \bdm-M by using two diffusion models, PVD \cite{zhou2021pvd} and \pc \cite{melas2023pc2}. Both the diffusion models use PVCNN as the backbone for predicting noise. Our Merging method focuses on the decoder part of PVCNN. To preserve the original knowledge in the reconstruction model, we freeze the encoder and finetune the decoder of \pc. Specifically, following \cite{zhang2023controlnet}, we retain the original encoders of both models while feeding the multi-scale features of PVD's encoder directly into {\pc}'s decoder. Each layer of the encoder enhanced by this integration is facilitated by a zero-initialized convolution layer. This setup enables a seamless and implicit merging of the knowledge from the prior model and the reconstruction model. 

\vspace{-1em}

\subsubsection{BDM-B (Blending)} Beyond the implicit incorporation of prior knowledge, we also introduce an explicit training-free fusion method on point clouds, termed as \bdm-B. It explicitly combines two groups of point clouds, $\bdmy^t=\{\mathbf{z}_{i}^t \}_{i=1}^N$ and $\bdmy_\gamma^t = \{ \mathbf{z}_{\gamma,i}^t \}_{i=1}^N$, each consisting of \(N\) points. These point clouds are noise-reduced versions derived from the same origin, generated by generation and reconstruction models separately. The blending operation employs a probabilistic function $\Psi$, which assigns a selection probability to each pair of corresponding points $\mathbf{z}_{j}^t, \mathbf{z}_{\gamma,j}^t$, where $\mathbf{z}_i^t$ denotes the \(i\)-th point in a point cloud $\bdmy$. The blending equation is defined as follows:
\begin{equation}
\bdmy_\pi^t = \Psi (\bdmy^t, \bdmy_\gamma^t)
\end{equation}

Additionally, assuming points are i.i.d., the whole point clouds can be formulated as 
\begin{equation}
p\left(\bdmy^t \mid \bdmy^{t+1}\right)=\prod_{i=1}^N p\left(\mathbf{z}_i^t \mid \mathbf{z}_i^{t+1}\right)
\end{equation}
and 
\begin{equation}
p_\gamma\left(\bdmy^t \mid \bdmy^{t+1}, \bdmx \right)=\prod_{i=1}^N p_\gamma\left(\mathbf{z}_i^t \mid \mathbf{z}_i^{t+1}, \bdmx \right)
\end{equation}

\noindent Points are selected based on $\Psi$; for instance, we might choose 50\% of the points from the prior and 50\% from the reconstruction model. Statistically, this approach is akin to \textbf{blending} two point clouds, resulting in a mixed distribution of points from both sources. The blending method also supports our Bayesian Diffusion Model theory, the details of which will be given in the Appendix.

\section{Experiment}
\begin{figure*}[ht]
\centering
\includegraphics[width=1.0\textwidth]{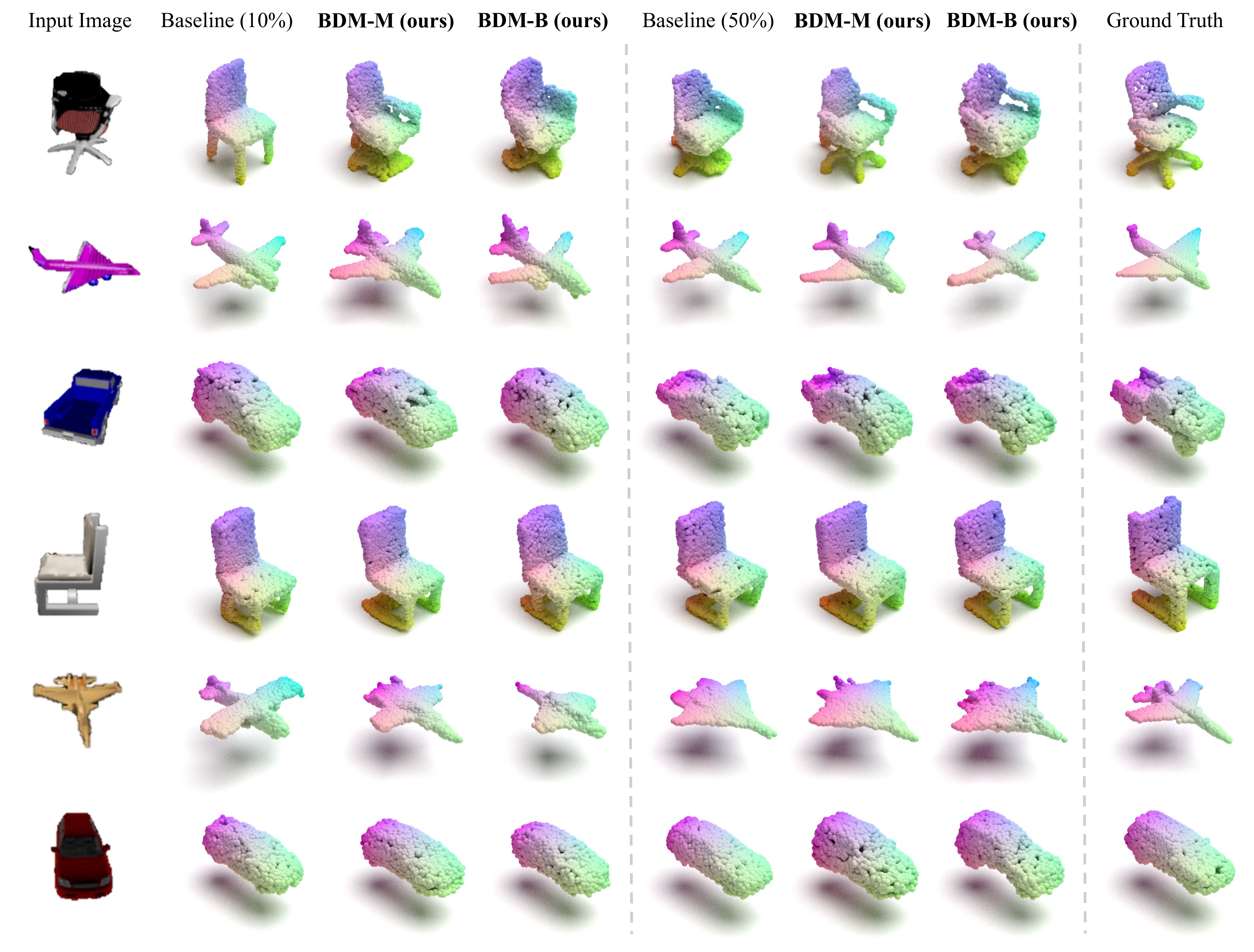}
\caption{Qualitative comparisons on the synthetic ShapeNet-R2N2 dataset. We use \pc \cite{melas2023pc2} and \ccd \cite{di2023ccd} as baselines of 3D shape reconstruction. Rows 1-3 show the visualization of \pc while rows 4-6 display the result of \ccd. We show our BDM's results under 10\% data in column 2-4 and the results under 50\% in column 5-7. Column 8 gives the corresponding ground truth.}
\label{fig:qualitative_shapenet}
\end{figure*}

\noindent \textbf{Dataset.} To demonstrate the efficacy of our Bayesian Diffusion Model, we conducted our experiments on two datasets: the synthetic dataset ShapeNet \cite{choy2016r2n2} and the real-world dataset Pix3D \cite{pix3d}. ShapeNet \cite{chang2015shapenet}, a collection of 3D CAD models, encompasses 3,315 categories from the WordNet database. Following prior work \cite{xie2020pix2vox++, melas2023pc2, di2023ccd}, we utilized a subset of three ShapeNet categories -- \textit{\{chair, airplane, car\}} -- from 3D-R2N2 \cite{choy2016r2n2}, including image renderings, camera matrices, and train-test splits. Pix3D comprises diverse real-world image-shape pairs with meticulously annotated 2D-3D alignments. For a balanced comparison with \ccd \cite{di2023ccd} and \pc \cite{melas2023pc2}, we reproduced these two works and adhered to \ccd's benchmark on three categories: \textit{\{chair, table, sofa\}}, allocating 80\% of samples for training and the remaining 20\% for testing. Further details about extended object categories are discussed in the Appendix.

\vspace{0.5em}
\noindent \textbf{Implementation Details.} For both the ShapeNet-R2N2 and Pix3D datasets, we sample 4,096 points per 3D object and set the rendering resolution to 224$\times$224. Notably, for Pix3D, images are cropped using their bounding boxes, necessitating adjustments to the camera matrices to accommodate the non-object-centric nature and varying sizes of the images. For the training of the generative diffusion model, we employ the PVD \cite{zhou2021pvd} architecture, adhering to their training methodologies. For the training of the reconstruction diffusion model, we select \pc and \ccd as two baselines and follow the recipe of \ccd. The inference step is set as 1,000 for both the generative model and the reconstruction model. In our \bdm inference framework, Bayesian integration is strategically applied at specific intervals during the denoising process. This integration occurs every 32 steps, both in the early stage and in the late stage of denoising. The fusion process initiated by this integration extends throughout 16 steps, ensuring a balanced and effective incorporation of Bayesian principles throughout the denoising procedure. Training of generative models was conducted on 4 NVIDIA A5000 GPUs, while we trained reconstruction models utilizing a single NVIDIA A5000 GPU. More details are available in the Appendix. 

\subsection{Quantitative Results}

We evaluate the performance of reconstruction with two widely recognized metrics: Chamfer Distance (CD) and F-Score@0.01 (F1). Chamfer Distance measures the disparity between two point sets by calculating the shortest distance from every point in one set to the closest point in the other set. To address the issue of CD's susceptibility to outliers, we additionally present F-Score at a threshold of 0.01. In this metric, a reconstructed point is deemed accurately predicted if its nearest distance to the points in the ground truth point cloud is within the specified threshold, which presents a measure of precision in the reconstruction process. 

\begin{figure*}[t]
\centering
\vspace{1em}
\includegraphics[width=1.0\textwidth]{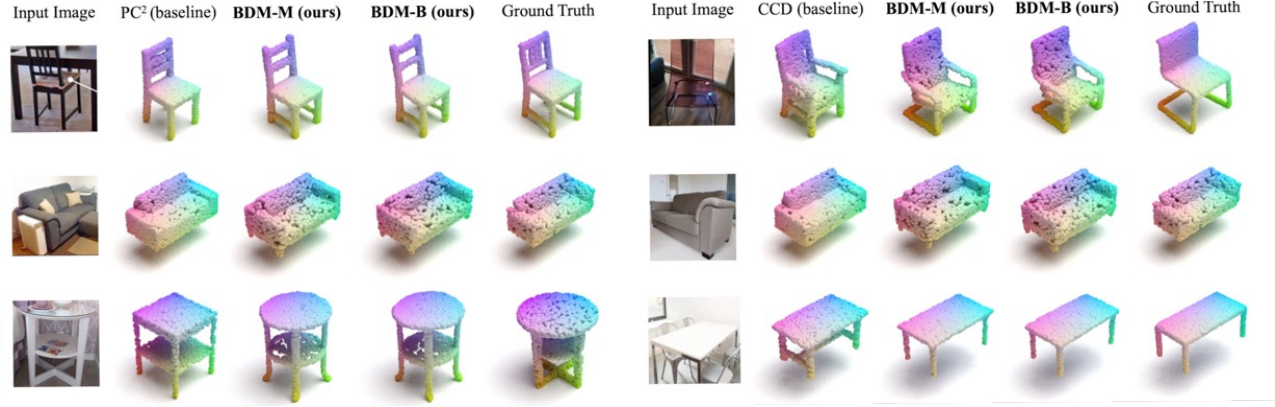}
\caption{Qualitative comparisons on the real-world Pix3D dataset. We examine three distinct categories, each represented in a separate row. Columns 3,4 and 8,9 feature our BDM, implemented based on \pc and \ccd respectively, for a comparative analysis.}
\label{fig:qualitative_pix3d}
\end{figure*}

\begin{table*}[htbp]
\centering
\resizebox{2.08\columnwidth}{!}
{
    \begin{tabular}{l||cc|cc|cc||cc|cc|cc||cc|cc|cc}
    \toprule
    \multicolumn{1}{c||}{} & \multicolumn{6}{c||}{\textbf{Chair}} & \multicolumn{6}{c||}{\textbf{Airplane}} & \multicolumn{6}{c}{\textbf{Car}}  \\
    \cmidrule{2-19}
      \textbf{Method} & \multicolumn{2}{c}{10\%} &\multicolumn{2}{c}{50\%} & \multicolumn{2}{c||}{100\%} & \multicolumn{2}{c}{10\%} & \multicolumn{2}{c}{50\%} & \multicolumn{2}{c||}{100\%} & \multicolumn{2}{c}{10\%} & \multicolumn{2}{c}{50\%} & \multicolumn{2}{c}{100\%} \\
    \cmidrule{2-19}
       & CD\down & F1\up & CD\down & F1\up & CD\down & F1\up & CD\down & F1\up & CD\down & F1\up & CD\down & F1\up & CD\down & F1\up & CD\down & F1\up & CD\down & F1\up \\ 
    \midrule 
    {\bf Base}: \pc \cite{melas2023pc2} & 97.25 & 0.393 & 73.58 & 0.437 & 65.57 & 0.464 & 88.00 & 0.605 & 76.39 & 0.628 & 65.97 & 0.655 & 64.99 & 0.524 & 62.59 & 0.542 & 64.36 & 0.547 \\
    \; BDM-M (ours) & 94.94 & 0.395 & 71.56 & 0.446 & 64.48 & 0.468 & 87.75 & 0.604 & 73.19 & 0.629 & 65.16 & 0.653 & 63.53 & 0.524 & {\bf 60.71} & 0.549 & {\bf 64.16} & 0.554 \\
    \; BDM-B (ours) & {\bf 94.67} & {\bf 0.410} & {\bf 69.99} & {\bf 0.463} & {\bf 64.21} & {\bf 0.485} & {\bf 83.62} & {\bf 0.612} & {\bf 68.66} & {\bf 0.641} & {\bf 59.04 }& {\bf 0.660} & {\bf 60.48} & {\bf 0.539} & 62.58 & {\bf 0.554} & 65.85 & {\bf 0.559} \\
    \midrule
    {\bf Base}: \ccd \cite{di2023ccd} & 89.79 & 0.418 & 63.13 & 0.474 & 58.47 & 0.498 & 81.29 & 0.612 & 72.46 & 0.635 & 62.77 & 0.651 & 63.13 & 0.531 & 62.25 & 0.550 & \cellcolorlightgray {\bf 61.88} & 0.562 \\
    \; BDM-M (ours) & 81.47 & 0.425 & 62.07 & 0.477 & 57.31 & 0.493 & 81.54 & 0.608	& 69.31 & 0.632	& 61.87 & 0.652 & 61.92 & 0.531 & 61.24 & 0.555 & 63.66 & 0.561 \\
    \; BDM-B (ours) & \cellcolorlightgray {\bf 79.26} & \cellcolorlightgray {\bf 0.441} & \cellcolorlightgray {\bf 60.07} & \cellcolorlightgray {\bf 0.497} & \cellcolorlightgray {\bf 56.78} & \cellcolorlightgray {\bf 0.510} & \cellcolorlightgray {\bf 77.34} & \cellcolorlightgray {\bf 0.621} & \cellcolorlightgray {\bf 66.83} & \cellcolorlightgray {\bf 0.644} & \cellcolorlightgray {\bf 56.96} & \cellcolorlightgray {\bf 0.660} & \cellcolorlightgray {\bf 59.05} & \cellcolorlightgray {\bf 0.546} & \cellcolorlightgray {\bf 60.59} & \cellcolorlightgray {\bf 0.560} & 66.90 & \cellcolorlightgray {\bf 0.569}\\
    \bottomrule
    \end{tabular}
}
\vspace{-0.5em}
\caption{Performance on \textit{Chair, Airplane} and \textit{Car} of ShapeNet-R2N2. We evaluate our BDM, comparing with two baselines: \pc and \ccd. These experiments span three different scales of training data (10\% / 50\% / 100\%) for the reconstruction diffusion model, demonstrating the efficacy of our \bdm method.}
\label{tab:quantitative_r2n2}
\end{table*}

\noindent \textbf{ShapeNet-R2N2.} \cref{tab:quantitative_r2n2} presents our BDM's performance across various training data scales on ShapeNet-R2N2. The results indicate improvement in both CD and F1 across all three categories. Notably, the improvement becomes more pronounced when training data scale decreases, highlighting the effectiveness of the prior facing data scarcity. 

\noindent \textbf{Pix3D.} Following \ccd, we also evaluate on the Pix3D dataset in \cref{tab:quantitative_pix3d}. It can be seen that our method effectively improves the performance and achieves state-of-the-art. Notably, we train the reconstruction model trained on Pix3D, while the prior model is trained on the same category of the separate, larger dataset: ShapeNet-R2N2 ($\sim$10 times size of Pix3D). In this way, we avoid the possible leaking and memorization problem, i.e., the generative model naively retrieve the nearest memorised shape learned from the reconstruction model. 

\begin{table}[!htbp]
\centering
\resizebox{\linewidth}{!}
{
    \begin{tabular}{l|cc|cc|cc}
    \toprule
    \multicolumn{1}{c|}{} & \multicolumn{2}{c|}{\textbf{Chair}} & \multicolumn{2}{c|}{\textbf{Sofa}} & \multicolumn{2}{c}{\textbf{Table}}  \\
    \multirow{-2}{*}{\textbf{Method}}  & CD\down & F1\up & CD\down & F1\up & CD\down & F1\up \\
    \midrule
    {\bf Base}: \pc \cite{melas2023pc2} & 115.94 & 0.443 & 47.17 & 0.445 & 202.77 & 0.397 \\
    \; BDM-M (ours) & 113.40 & 0.449 & \cellcolorlightgray {\bf 44.50} & 0.451 & 202.08 & 0.413  \\
    \; BDM-B (ours) & {\bf 110.60} & {\bf 0.455} & 45.05 & {\bf 0.455} & \cellcolorlightgray {\bf 186.46} & \cellcolorlightgray {\bf 0.429} \\
    \midrule
    {\bf Base}: \ccd \cite{di2023ccd} & 111.42 & 0.456 & 44.91 & 0.450 & 196.28 & 0.418  \\
    \; BDM-M (ours) & \cellcolorlightgray {\bf 110.86} & 0.464 & 44.86 & 0.452 & {\bf 193.56} & {\bf 0.424} \\
    \; BDM-B (ours) & 111.12 & \cellcolorlightgray {\bf 0.466} & {\bf 44.51} & \cellcolorlightgray {\bf 0.456} & 194.91 & 0.423 \\
    \bottomrule
    \end{tabular}
}
\vspace{-0.5em}
\caption{Performance on \textit{Chair, Sofa} and \textit{Table} of Pix3D. We evaluate our BDM on the two baselines: \pc and \ccd.} 
\label{tab:quantitative_pix3d}
\vspace{-2em}
\end{table}

\vspace{-0.5em}
\subsection{Qualitative Results}
In addition to the quantitative study, we also show the qualitative results to show the superiority of our \bdm on both ShapeNet and Pix3D. For the ShapeNet-R2N2 dataset, \cref{fig:qualitative_shapenet} demonstrates our \bdm's capacity in synthetic 3D object reconstruction. In \cref{fig:qualitative_pix3d}, it can be seen clearly that our method surpasses baselines with respect to the reconstruction quality on the Pix3D dataset. 
Notably, our \bdm effectively restores the missing spindles of the chair in the first image and eliminates hallucinations in the last image. 

\vspace{-0.5em}
\subsection{Efficiency and Fairness Analysis}
As shown in \cref{tab:efficiency}, we present \bdm's parameters, runtime and GPU memory when doing inference with batch size of 1. While parameters increase due to the incorporation of prior model $\mathcal{P}$, memory usage and runtime of \bdm only increase slightly. Moreover, \bdm leverages off-the-shelf pre-trained weights for prior model $\mathcal{P}$ which is \textbf{frozen}, which doesn't further incur additional training cost. 

\begin{table}[hbt]
\centering
\vspace{-0.5em}
\resizebox{\linewidth}{!}{
    \begin{tabular}{c|c|c|c}
    \toprule
    & \textbf{PC$^2$/CCD-3DR} & \textbf{BDM-B} & \textbf{BDM-M} \\
    \midrule
    \textbf{\#Parameters (M)} & 47.41 & 73.78 & 74.82 \\
    \textbf{Runtime (s)} & 46.89 & 48.84 & 49.24 \\
    \textbf{GPU memory (GB)} & 1.73 & 1.93 & 2.01 \\
    \bottomrule
    \end{tabular}
}
\vspace{-0.5em}
\caption{Model parameters and inference-time efficiency.}
\vspace{-1.5em}
\label{tab:efficiency}
\end{table}

\subsection{Ablation Study}

To validate the effectiveness and rationality of our \bdm, we conduct several ablation studies to explore the impact of the timing, duration and intensity to absorb priors. Unless otherwise specified, we set BDM-B comprising PVD trained on 100\% data and \ccd trained on 10\% data from ShapeNet-\textit{chair} as our baseline.

\noindent \textbf{Prior Integration Timing.} This subsection evaluates the effectiveness of integrating prior knowledge at various stages of the denoising process. \cref{tab:prior_time} reveals that integrating priors in the late stage alone yields significant improvement on model performance, reducing CD to 80.22 and increasing F1 to 0.436. Furthermore, combining early and late-stage integration further enhances results, achieving the lowest CD of 79.26 and the highest F1 of 0.441. This contrasts with the middle stage integration, which even degrades the performance of the baseline on F1. 

\begin{table}[!htp]
\centering
\vspace{-0.5em}
\resizebox{\linewidth}{!}{
\begin{tabular}{ccc|cc}
\toprule
\textbf{Early} & \textbf{Middle} & \textbf{Late} & \textbf{Chamfer Distance} \down & \textbf{F-Score@0.01} \up \\
\midrule
& & & 89.79 & 0.418 \\
\cmark & & & 82.34 & 0.421 \\
& \cmark & & 87.61 & 0.395 \\
& & \cmark & 80.22 & 0.436 \\
\cmark & & \cmark & \cellcolorlightgray \textbf{79.26} & \cellcolorlightgray \textbf{0.441} \\
\cmark & \cmark & \cmark & 81.92 & 0.416 \\
\bottomrule
\end{tabular}
}
\vspace{-0.5em}
\caption{Ablation on the timing of prior integration. This table presents the impact of applying prior integration during the early, middle, and late stages of the denoising process on CD and F1.}
\vspace{-1em}
\label{tab:prior_time}
\end{table}

\noindent \textbf{Prior Integration Duration.} As shown in \cref{tab:prior_duration}, we investigate how varying the duration of prior integration affects the denoising process. First, the performance will greatly improve once the prior duration is greater than 1, thereby strongly validating the effectiveness of our \bdm. Notably, the prior integration duration of 16 steps demonstrates the most substantial improvement. This is a remarkable improvement over the baseline (0 step). The diminishing returns are observed with the duration of 32 steps, suggesting that a moderate duration of prior integration optimally balances denoising effectiveness and prior guidance. 

\begin{table}[!ht]
\centering
\vspace{-0.5em}
\resizebox{\linewidth}{!}{
    \begin{tabular}{l|ccccccc}
    \toprule
    \textbf{Prior Duration} & \textbf{0 step (baseline)} & \textbf{1 step} & \textbf{2 step} & \textbf{4 step} & \textbf{8 step} & \textbf{16 step} & \textbf{32 step} \\
    \midrule
    \textbf{Chamfer Distance} \down & 89.79 & 80.89 & 81.02 & 80.65 & 79.72 & \cellcolorlightgray \textbf{79.26} & 79.94 \\
    \textbf{F-Score@0.01} \up & 0.418 & 0.427 & 0.430 & 0.429 & 0.432 & \cellcolorlightgray \textbf{0.441} & 0.438 \\
    \bottomrule
    \end{tabular}
}
\vspace{-0.5em}
\caption{Ablation on the duration of prior integration. This table presents how different durations of prior integration affect CD and F1, ranging from 0 to 32 steps.}
\vspace{-1em}
\label{tab:prior_duration}
\end{table}

\noindent \textbf{Prior Integration Ratio.} In this part, we present an ablation on the impact of the prior integration ratio on the BDM-B process. As can be seen in \cref{tab:prior_ratio}, interestingly, a gradual increase in the prior take-in ratio yields varying results. At 25\%, there is a minor detriment to performance. However, at 50\%, we observe a significant improvement, marking the optimal balance in prior integration. On the contrary, further increasing the ratio to 75\% and 100\% leads to a drastic decline in performance. These results suggest that while a moderate level of prior integration enhances the model’s performance, excessive integration can be detrimental.

\begin{table}[h]
\centering
\vspace{-0.5em}
\resizebox{\linewidth}{!}{
\begin{tabular}{l|ccccc}
\toprule
\textbf{Prior Ratio} & \textbf{0\% (baseline)} & \textbf{25\%} & \textbf{50\%} & \textbf{75\%} & \textbf{100\%} \\
\midrule
\textbf{Chamfer Distance} \down & 89.79 & 88.22 & \cellcolorlightgray \textbf{79.26} & 142.69 & 256.13 \\
\textbf{F-Score@0.01} \up & 0.418 & 0.382 & \cellcolorlightgray \textbf{0.441} & 0.307 & 0.242 \\
\bottomrule
\end{tabular}
}
\vspace{-0.5em}
\caption{Ablation on the ratio of prior integration. This table compares the effects of different prior integration ratios on CD and F1.}
\vspace{-1.5em}
\label{tab:prior_ratio}
\end{table}

\subsection{BDM vs CFG}

Considering that Classifier-Free Guidance (CFG) \cite{ho2022cfg} also shows the relation in the inference stage between the gradient $\varepsilon(\bdmy^{t})$ from the unconditional diffusion model and the gradient $\varepsilon_{\gamma}(\bdmy^{t},\bdmx)$ from the conditional diffusion model, we conduct an ablation study to compare our method with CFG. As shown in \cref{tab:compare_with_cfg}, both our methods surpass CFG. 

\begin{table}[h]
\centering
\vspace{-0.5em}
\resizebox{0.9\linewidth}{!}
{
    \begin{tabular}{l|cc}
    \toprule
     & Chamfer Distance \down & F-Score@0.01 \up \\
    \midrule
    Baseline & 58.47 & 0.498 \\
    CFG \cite{ho2022cfg} & 59.08 & 0.495 \\
    \bdm & \cellcolorlightgray \textbf{56.78} & \cellcolorlightgray \textbf{0.510} \\
    \bottomrule
    \end{tabular}
}
\vspace{-0.5em}
\caption{We compare our BDM with CFG. We train these models and test these two methods on 100\% data of chair from ShapeNet.}
\vspace{-1.5em}
\label{tab:compare_with_cfg}
\end{table}

\vspace{-0.5em}
\subsection{Human Evaluation}

To better evaluate the reconstruction quality, we also conducted human evaluation. We randomly selected 20 comparison groups from the Chair, Airplane, and Car classes in the ShapeNet dataset, totaling 60 groups. In each group, we present outputs generated by CCD-3DR, BDM-B, and BDM-M. Sixteen evaluators then ranked each group on a scale of 1 to 3, and the average scores are shown in \cref{tab:human}. The results show that our two methods, BDM-M and BDM-B, still outperforms CCD-3DR, which is aligned with the quantitative result presented in the main paper. More details will be discussed in the Appendix. 

\begin{table}[htbp]
\centering
\vspace{-0.5em}
\resizebox{.9\linewidth}{!}{
    \begin{tabular}{l|ccc}
    \toprule
    Human Evaluation & CCD & BDM-B & BDM-M \\
    \midrule
    Chair & 1.48 & 2.15 & 2.32 \\
    Airplane & 1.74 & 1.86 & 2.40 \\
    Car & 1.77 & 1.88 & 2.35 \\
    \midrule
    Average      & 1.67 & 1.96 & 2.35 \\
    \bottomrule
    \end{tabular}
}
\vspace{-0.5em}
\caption{Human Evaluation over 3 categories for CCD, BDM-B, and BDM-M, with 3 being the best and 1 being the worst. }
\vspace{-2em}
\label{tab:human}
\end{table}

\section{Conclusion and Limitations} 
\vspace{-0.5em}

In this paper, we present Bayesian Diffusion Model (BDM), a novel diffusion-based inference method for posterior estimation. BDM overcomes the limitations in the traditional MCMC-based Bayesian inference that requires having the explicit distributions in performing stochastic gradient Langevin dynamics by tightly coupling the bottom-up and top-down diffusion processes using learned gradient computation networks. We show a plug-and-play version of the BDM (BDM-B) and a learned fusion version (BDM-M). BDM is particularly effective for applications where paired data-labels such as image-object point clouds, are scares, while standalone labels, like object point clouds, are abundant. It demonstrates the state-of-the-art results for single image 3D shape reconstruction. BDM points to a promising direction to perform the general inference in computer vision and machine learning beyond the 3D shape reconstruction application shown here. 

\noindent \textbf{Limitations}. BDM requires both the prior and data-driven processes to be diffusion processes. Also, BDM-B takes advantage of the explicit representation of point clouds, which might not be broadly adopted on implicit representations. 

\noindent \textbf{Acknowledgement.} This work is supported by NSF Award IIS-2127544. We are grateful for the constructive feedback by Zirui Wang and Zheng Ding. 

{
    \small
    \bibliographystyle{ieeenat_fullname}
    \bibliography{main}
}

\cleardoublepage
\newpage
\twocolumn[
    \centering
    \Large
    \textbf{Appendix} \\
    \vspace{1.0em}
]

\crefname{section}{Appendix}{Appendices}
\appendix

\renewcommand{\thepart}{\Alph{part}}
\stepcounter{part}
\setcounter{figure}{0}
\setcounter{table}{0}
\setcounter{equation}{0}
\counterwithin{figure}{part}
\counterwithin{table}{part}
\counterwithin{equation}{part}

\section{Training and Inference Details}
\label{appendix:train_inference_detail}

We train the generative model PVD~\cite{zhou2021pvd} with a batch size of 128 for 10k iterations, and adopt Adam optimizer with learning rate $2\times10^{-4}$. For \textit{airplane}, we set $\beta_0 = 10^{-5}$ and $\beta_T = 0.008$. For other categories, we set $\beta_0 = 10^{-4}$ and $\beta_T = 0.01$. 

In terms of reconstruction model \pc / \ccd, we train it with a batch size of 16 for 100k iterations, and adopt Adam optimizer with a learning rate increasing from $10^{-5}$ to $10^{-3}$ in the first 2k iterations then decaying to 0 in the remaining iterations. For all the categories, we set $\beta_0 = 10^{-5}$ and $\beta_T = 0.008$. 

For the training of \bdm-merging, we adopt a similar strategy as the reconstruction model, except that we reduce the total iterations to 20k and scale the learning rate schedule accordingly.

During inference, we set the number of denoising steps as 1000. We divide the denoising process into three distinct stages, \ie early (timesteps 1000–872), middle (timesteps 872–128), and late (timesteps 128–0). We conduct our Bayesian denoising steps in the early and late stages. To be more specific, every 32 timesteps, a Bayesian denoising step is executed for a duration of 16 timesteps. Subsequently, we forward a standard reconstruction process for 16 timesteps, followed by another Bayesian denoising step.

\section{Extended Object Categories}
\label{appendix:extendided_obj_categories}
Leveraging PVD as our prior model, we follow its settings and adopt official pre-trained weights, which are only on three categories. Each category is trained with different hyperparameters. To further illustrate the effectiveness of \bdm, we add two new categories of ShapeNet-R2N2 in \cref{tab:additional_category}.

\begin{table}[h]
\centering
\resizebox{\linewidth}{!}
{
    \begin{tabular}{l|cc|cc|cc|cc|cc|cc}
    \toprule
    \multicolumn{1}{c|}{} & \multicolumn{6}{c|}{\textbf{Sofa}} & \multicolumn{6}{c}{\textbf{Table}}  \\
    \cmidrule{2-13}
      \textbf{Method} & \multicolumn{2}{c}{10\%} &\multicolumn{2}{c}{50\%} & \multicolumn{2}{c|}{100\%} & \multicolumn{2}{c}{10\%} & \multicolumn{2}{c}{50\%} & \multicolumn{2}{c}{100\%} \\
    \cmidrule{2-13}
       & CD\down & F1\up & CD\down & F1\up & CD\down & F1\up & CD\down & F1\up & CD\down & F1\up & CD\down & F1\up \\ 
    \midrule 
    CCD-3DR & 63.20 & 0.444 & 47.83 & 0.482 & 43.16 & 0.501 & 79.41 & 0.523 & 77.51 & 0.520 & 67.13 & 0.538\\
    \bdm-merging & 62.29 & 0.460 & 45.18 & 0.500 & 41.43 & 0.517 & 78.25 & 0.535 & 75.94 & \cellcolorlightgray \textbf{0.538} & 65.24 & 
    \cellcolorlightgray \textbf{0.560} \\
    \bdm-blending & 
    \cellcolorlightgray \textbf{61.54} & 
    \cellcolorlightgray \textbf{0.471} & 
    \cellcolorlightgray \textbf{44.31} & 
    \cellcolorlightgray \textbf{0.516} & 
    \cellcolorlightgray \textbf{41.94} & 
    \cellcolorlightgray \textbf{0.520} & 
    \cellcolorlightgray \textbf{74.18} & 
    \cellcolorlightgray \textbf{0.547} & 
    \cellcolorlightgray \textbf{73.46} & 
     0.526 & 
    \cellcolorlightgray \textbf{64.56} & 
    0.557 \\
    \bottomrule
    \end{tabular}
}
\vspace{-0.5em}
\caption{Results on two additional categories of ShapeNet-R2N2, \ie \textit{sofa} and \textit{table}.}
\vspace{-2em}
\label{tab:additional_category}
\end{table}

\section{Alternative Prior Model}
\label{appendix:prior_select}

\begin{figure}[h]
\centering
\includegraphics[width=\linewidth]{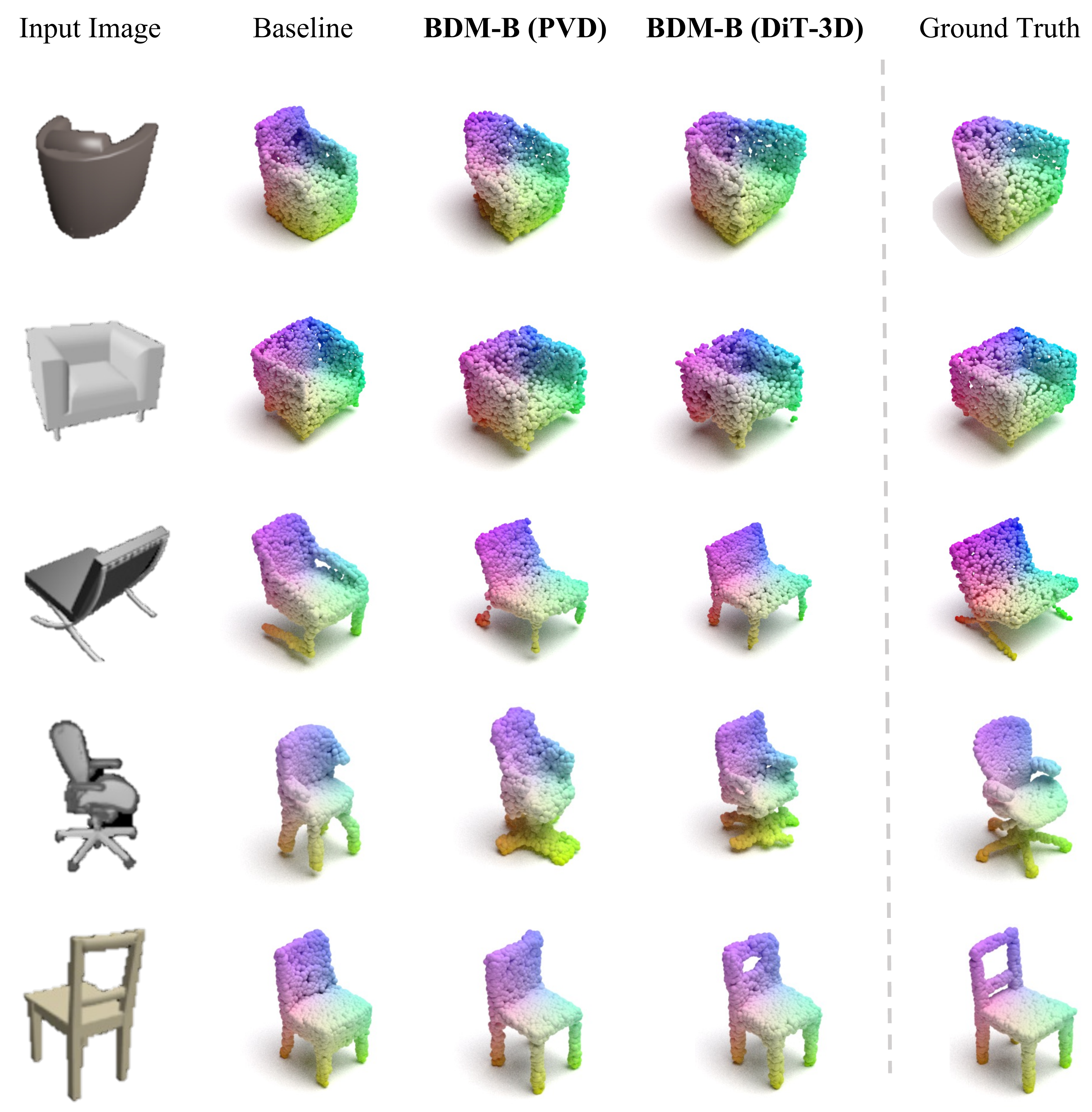}
\vspace{-0.5em}
\caption{Visualization of taking DiT-3D as prior compared with PVD on chairs.}
\vspace{-1.5em}
\label{fig:dit3d_vis}
\end{figure}

 To validate the robustness of our proposed \bdm, we explore its performance with alternative generative priors. Specifically, we replace the PVD-based generative diffusion model~\cite{zhou2021pvd} with DiT-3D~\cite{mo2023dit3d}, an extension of DiT~\cite{Peebles2022DiT}, which uniquely applies the denoising process to voxelized point clouds. Unlike the PVCNN architecture~\cite{liu2019pvcnn}, DiT-3D leverages a Transformer-based framework, and therefore we only experiment on \bdm-blending. For proof-of-concept purposes, we take 10\% of the chairs on ShapeNet as the training data for the reconstruction model (\ccd). As demonstrated in \cref{tab:dit3d}, \bdm brings consistent improvement regardless of the prior model utilized. Also, we show some qualitative visualizations in \cref{fig:dit3d_vis}. The results illustrate the effectiveness of our \bdm across different generative diffusion priors.

 \begin{table}[htb]
\centering
\vspace{-0.5em}
\resizebox{.6\linewidth}{!}{
\begin{tabular}{l|cc}
\toprule
 & CD \down & F1 \up \\
\midrule
CCD-3DR (baseline) & 89.79 & 0.418 \\
+ PVD \cite{zhou2021pvd} & 79.26 & 0.441 \\
+ DiT-3D \cite{mo2023dit3d} & 80.77 & 0.431 \\
\bottomrule
\end{tabular}
}
\vspace{-0.5em}
\caption{Reconstruction results of taking DiT-3D as prior compared with PVD, evaluated with Chamfer Distance and F-Score@0.01.}
\vspace{-1.5em}
\label{tab:dit3d}
\end{table}

\section{Different Gaussian Noises}
\label{appendix:different_noise}

\begin{figure}[htb]
    \centering
    \vspace{-1em}
    \includegraphics[width=\linewidth]{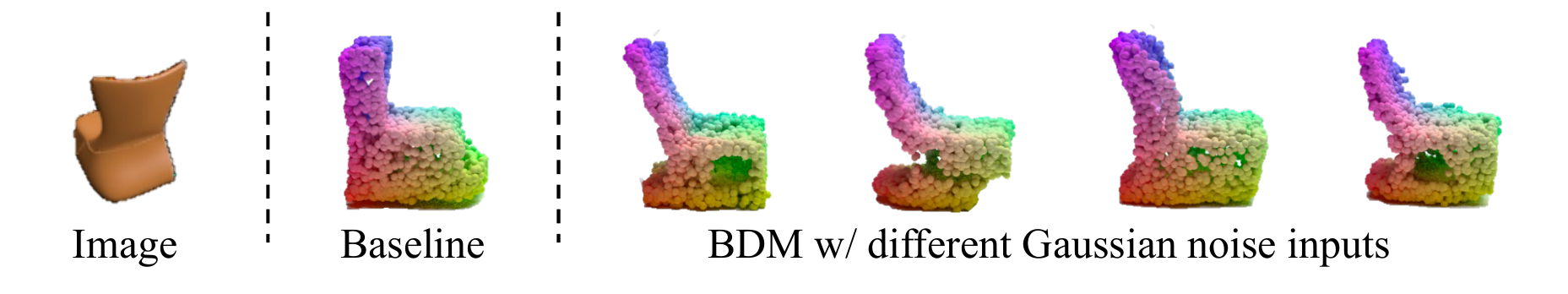}
    \vspace{-2em}
    \caption{Predictions over different initial Gaussian noise.}
    \vspace{-0.5em}
    \label{fig:diff_gaussian_noise}
\end{figure}

One key advantage of Bayesian methods is that it allows to obtain distributions over predicted outputs, allowing to measure prediction uncertainties. The uncertainties in diffusion model originate from initial Gaussian noise. To investigate the correlation between the outputs and different initial Gaussian noise $\mathbf{X}_T$, we evaluate \bdm-blending with 10 different initial noise inputs, and report mean and variance in \cref{tab:mean_var_initial_gaussian_noise}. In addition, we visualize a chair sample in \cref{fig:diff_gaussian_noise}, featuring consistent valid reconstructions across different noise inputs despite minor shape differences. \bdm transforms the vertical chair-back to a tilted and curved one, better aligned with the 2D image.

\begin{table}[htb]
\centering
\vspace{-0.5em}
\resizebox{0.6\linewidth}{!}{
    \begin{tabular}{l|cc}
    \toprule
     & CD $\downarrow$ & F1 $\uparrow$ \\
    \midrule
    CCD-3DR & 89.79 & 0.418\\
    \midrule
    $\mathbf{X}_T$ ablate & \textbf{79.61} (0.048) & \textbf{0.441} (2.2e-5) \\
    \bottomrule
    \end{tabular}
}
\vspace{-0.5em}
\caption{Mean and variance w.r.t. different initial Gaussian noise.}
\vspace{-1.5em}
\label{tab:mean_var_initial_gaussian_noise}
\end{table}

\section{Details of Human Evaluation}
\label{appendix:detail_human_eval}
\begin{figure}[h]
\centering
\includegraphics[width=0.9\linewidth,trim=0 4em 0 0,clip]{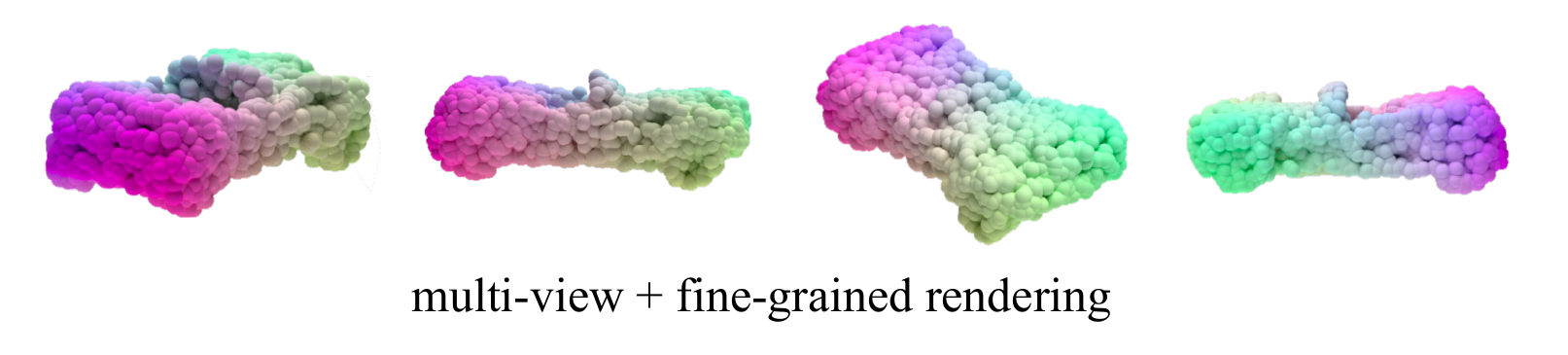}
\caption{Multi-view rendering.} 
\vspace{-1em}
\label{fig:vis_human_eval_cur}
\end{figure}

For each generated 3D point cloud, we render multiple images of it, as shown in \cref{fig:vis_human_eval_cur}. 

From the evaluation results, we can see our BDM-M and BDM-B both outperform CCD-3DR, while the quality of BDM-M is more favored than BDM-B. However, this superiority of BDM-M is not evident from the CD and F1. 

\begin{figure}[h]
    \centering
    \includegraphics[width=\linewidth,trim=0 1em 0 1pt,clip]{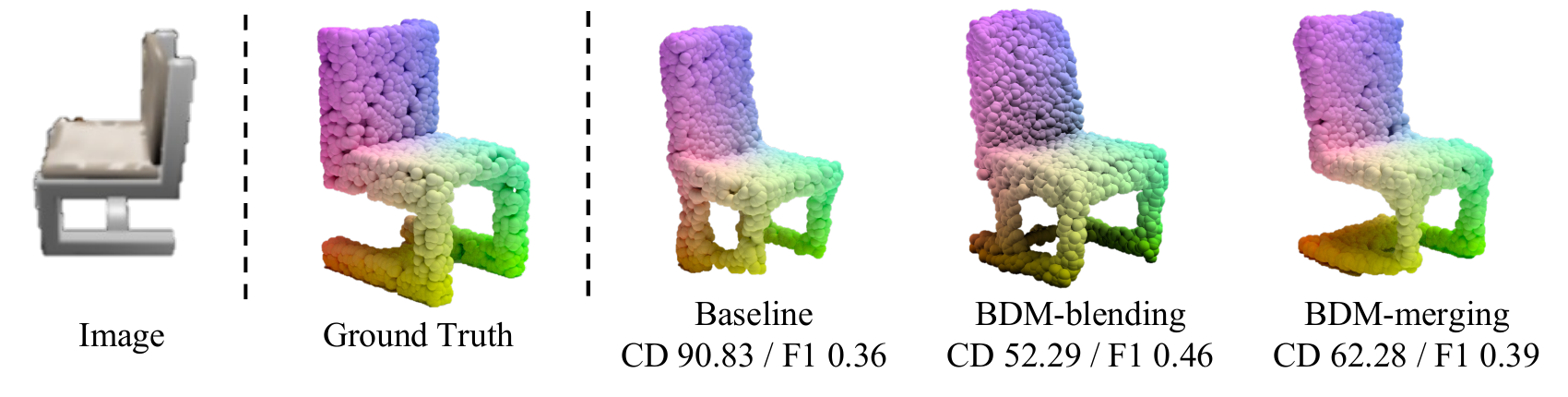}
    \vspace{-2em}
    \caption{Quantitative results vs actual visualization.}
    \label{fig:quan_vs_human_eval}
\end{figure}

As discussed in \cite{tatarchenko2019single, wu2021density}, CD is susceptible to mismatched local density and F1 does not fully address such issue. These metrics may not align with human preference, as shown in \cref{fig:quan_vs_human_eval}. Therefore, according to the human evaluation, BDM-M yields more visually appealing outcomes whereas blending has stronger quantitative results, which confirms the effectiveness of our BDM-M approach. 

\end{document}